\documentclass[10pt,twocolumn,letterpaper]{article}

\usepackage{cvpr}
\usepackage{times}
\usepackage{epsfig}
\usepackage{graphicx}
\usepackage{amsmath}
\usepackage{amssymb}

\usepackage{booktabs}
\usepackage{comment}
\usepackage{multirow}
\usepackage{algorithm,algpseudocode}
\usepackage{url}
\usepackage{overpic}
\usepackage[font=normalsize]{subfig}

\usepackage{bbding}
\usepackage{xspace}
\usepackage{tabularx}
\usepackage{verbatim}
\usepackage{wasysym} 
\usepackage{caption}
\usepackage{float}
\usepackage[font=small,tableposition=top]{caption}
\usepackage{lipsum}
\usepackage[table,x11names]{xcolor}
\usepackage{bm}
\definecolor{mygray}{gray}{0.95}

\newcommand{\taohu}[1]{\textcolor{purple}{[\textbf{Tao}: #1]}}
\newcommand{\wt}[1]{\textcolor{orange}{[\textbf{WT}: #1]}}

\newcommand{\tablestyle}[2]{\setlength{\tabcolsep}{#1}\renewcommand{\arraystretch}{#2}\centering\small}

\usepackage[pagebackref=true,breaklinks=true,letterpaper=true,colorlinks,bookmarks=false]{hyperref}

 \cvprfinalcopy 

\def\cvprPaperID{} 

\ifcvprfinal\fi
\begin{document}


\title{Query by Activity Video in the Wild}

\author{Tao HU \qquad William Thong \qquad Pascal Mettes \qquad Cees G.M. Snoek \\
University of Amsterdam\\
}

\maketitle


\begin{abstract}
This paper focuses on activity retrieval from a video query in an imbalanced scenario. In current query-by-activity-video literature, a common assumption is that all activities have sufficient labelled examples when learning an embedding. This assumption does however practically not hold, as only a portion of activities have many examples, while other activities are only described by few examples. In this paper, we propose a visual-semantic embedding network that explicitly deals with the imbalanced scenario for activity retrieval. Our network contains two novel modules. The visual alignment module performs a global alignment between the input video and fixed-sized visual bank representations for all activities. The semantic module performs an alignment between the input video and fixed-sized semantic activity representations. By matching videos with both visual and semantic activity representations that are of equal size over all activities, we no longer ignore infrequent activities during retrieval. Experiments on a new imbalanced activity retrieval benchmark show the effectiveness of our approach for all types of activities.
\end{abstract}

\section{Introduction}
This paper investigates the problem of activity retrieval given a video as example. In current literature, activity retrieval is more often framed as a classification task~\cite{kang2003query}, a localization task~\cite{anne2017localizing}, or as a retrieval-by-text problem~\cite{escorcia2019temporal}. For the less common task of activity retrieval by video, several works have shown that activities can be retrieved directly from a user-provided query~\cite{ciptadi2014movement,douze2016circulant,qin2017fast,snoek2009concept,song2018self}. A standard assumption however is that the activities form a closed set, \ie they assume a fixed set of activities, each with many training videos. In practice, most activities will have few examples. Without an explicit focus on such activities, they will be ignored in favour of activities with many examples. In this work, we focus on learning balanced video representations of activities for retrieval, regardless of whether they have many or few examples.

 \begin{figure}
    \centering
    \includegraphics[width=0.9\columnwidth]{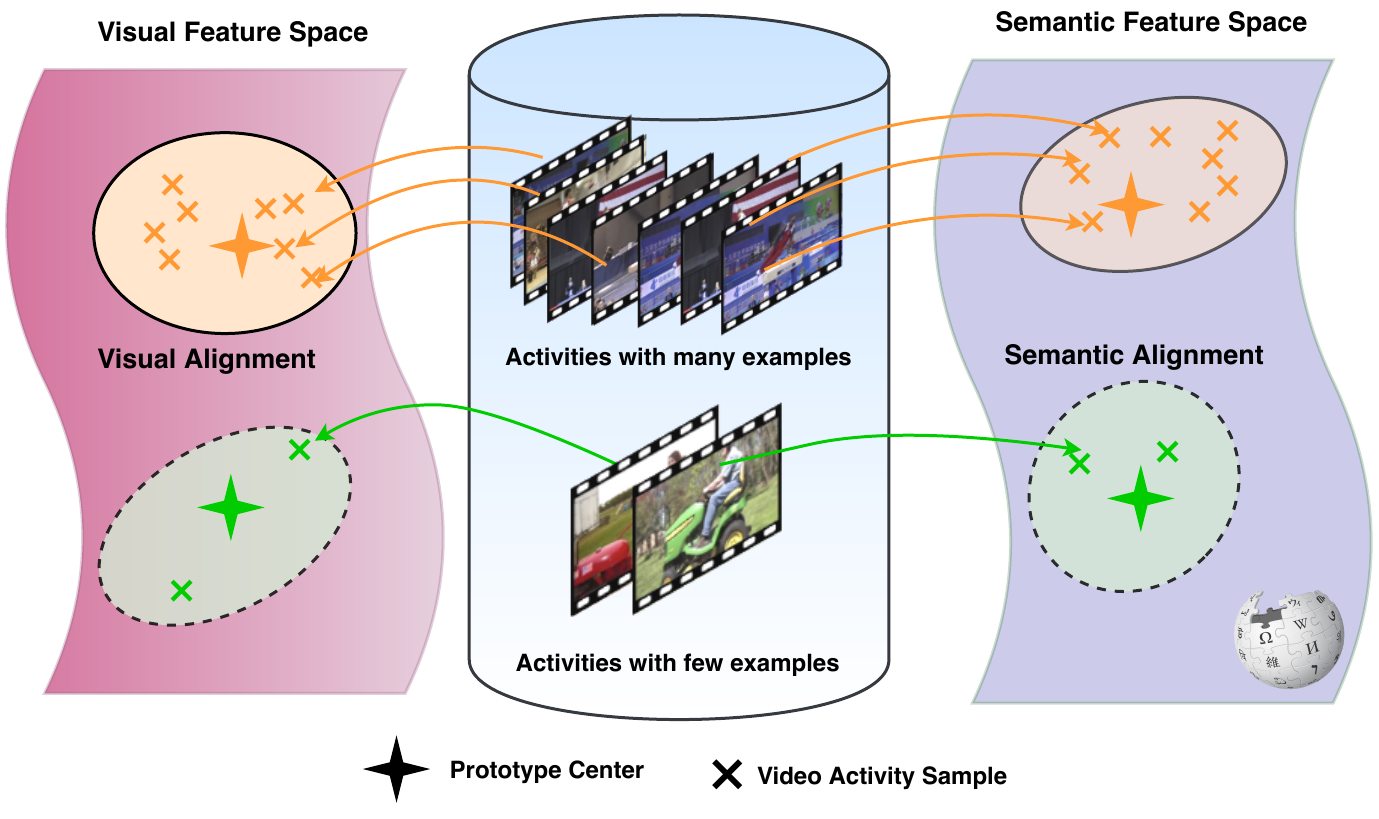}
    \caption{\textbf{Our motivation.}  We aim at retrieving activities by an activity video query. The training set is composed of activities with many examples and activities with few examples. We propose a visual-semantic alignment to balance the retrieval performance between base and novel classes.}
    \label{fig:motivation}
\end{figure}

Learning with imbalanced data is an active research topics for various visual tasks, including image classification~\cite{liu2019large,zhu2014capturing}, image segmentation~\cite{BuloNK17,kervadec19a}, and object detection~\cite{oksuz2019imbalance}. A central theme in these works is to either make classes with few examples more prominent, or switch to a setting where all classes have an equally-sized representation, \eg using memory banks~\cite{luo2019few,zhu2018compound} or prototypes~\cite{mettes2019hyperspherical,prototypical}. Here, we take inspiration from existing imbalanced tasks for the problem of imbalanced activity retrieval from video queries. We seek to introduce two alignment modules to match the activity feature regardless of whether they have many or few examples, see Figure~\ref{fig:motivation}. Different from current works, we do so by using both visual and semantic prototypes, where we emphasize the importance of a global alignment with respect to all activities. This allows us to better focus on equally-sized activity representations during training, which in turn results in a more balanced retrieval.

As a first contribution in this work, we introduce a new task about video query by activity in the wild and emphasize the importance of performance balance between the activities with many examples and activities with few examples. Second, we introduce a visual-semantic embedding network for retrieval by a video query. The network extends the standard classification loss in deep networks with two novel modules. The visual alignment module  maintains a visual bank with equal space for each activity. The representation of an input video is globally aligned with the visual bank representations of all activities to obtain a loss that disregards the amount of examples available for each activity. The semantic alignment module performs a similar global alignment and loss, but between the input video and video-independent semantic representations of activities such as word embeddings. These modules explicitly target the problem of imbalance in video dataset. Third, we reorganized the ActivityNet dataset~\cite{caba2015activitynet} to emulate an imbalanced retrieval dataset, along with new data splits and example sampling. 
we perform extensive evaluation and analyses to examine the workings of our approach for imbalanced activity retrieval. Lastly, we show our ability on video clips~\cite{miech2019howto100m} and moments~\cite{escorcia2019temporal}.

\section{Related work}

\subsection{Video retrieval}
For video retrieval, one common direction is to retrieve videos by a textual query~\cite{anne2017localizing,gao2017tall,hendricks2018localizing,miech2019howto100m,mithun2019weakly,otani2016learning,torabi2016learning,xu2019multilevel}. Hendricks \etal~\cite{anne2017localizing} propose a network that localizes text queries in videos using local and global temporal video representations. Hendricks \etal~\cite{hendricks2018localizing} further propose to model the context as a latent variable to bridge the gap between videos and textual queries. Beyond video retrieval, a number of recent works have investigated localized retrieval from text queries. Notably, Gao \etal~\cite{gao2017tall} and Miech \etal~\cite{miech2019howto100m} jointly model text and video clips in a shared space to obtain fixed-length local videos clips as retrieval output. Similar endeavours have been proposed to retrieve localized video moments from untrimmed datasets given a text query~\cite{mithun2019weakly}. In this work, we also extend the retrieval beyond videos only to clips and moments, but do so by using an input video as query, rather than text.

For activity retrieval by query video, current works are generally concerned with an efficient matching setup between query and test videos. Examples include retrieval using hashing~\cite{song2018self} and retrieval using quantized video representations~\cite{ciptadi2014movement}. A common starting assumptions is that the activities to retrieve have ample training examples to learn such an efficient matching. In this work, we challenge this assumption and propose a network is retrieve both activities with many examples and activities with few examples from a query video.

\subsection{Learning with imbalanced data}
When dealing with frequent classes (base classes~\cite{hariharan2017low}) and infrequent classes (novel classes~\cite{hariharan2017low}), a persistent issue is overfitting to the base classes. Transfer learning to novel classes provides a way to boost the performance on novel classes, although this is paired a catastrophic forgetting problem on base classes~\cite{serra2018overcoming}. Meta-learning is similarly focused on improving generalization to novel classes, \eg through a few steps of fine-tuning~\cite{maml,rusu2018meta}. However, these methods only consider the generalization on novel classes while ignoring the performance of base classes~\cite{luo2019few}. We aim to achieve a balance between both.

To attain such a balance, early work attempted to use a single example of the novel class to adapt classifiers from similar base classes using hand-crafted features~\cite{bart2005cross}. Learning with imbalanced classes has since actively been researched in image classification~\cite{liu2019large,zhu2014capturing}, image segmentation~\cite{BuloNK17,kervadec19a}, and object detection~\cite{oksuz2019imbalance}. Bharath~\etal~\cite{hariharan2017low} for example tackle the imbalance problem by hallucinating additional training examples for rare novel classes. As an extension of these works, we explore the balancing of base and novel classes for the problem of activity retrieval by a video example.

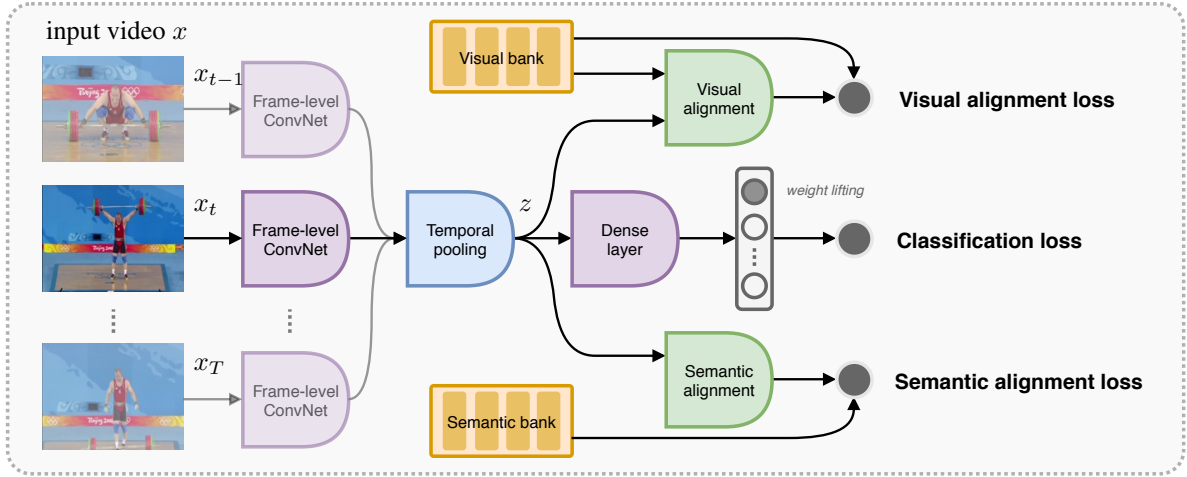
\begin{figure*}
    \centering
    \begin{overpic}[width=0.9\linewidth]{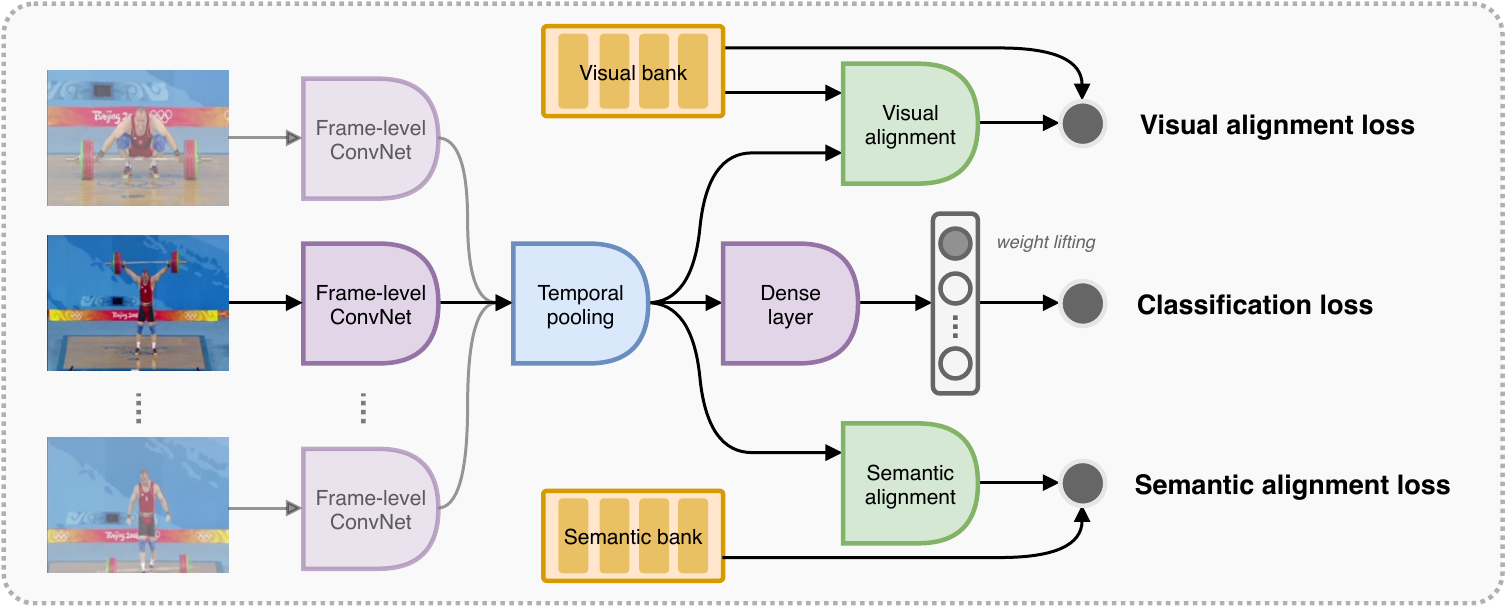}
    \put(16,33.5){$x_{t-1}$}
    \put(16,22.5){$x_t$}
    \put(16,9){$x_{T}$}
    \put(43.5,22.5){$z$}
    \put(3.5,37){input video $x$}
    \end{overpic}
    \caption{\textbf{Visual-Semantic Embedding Network}. For training phase, we adopt the visual alignment loss, semantic alignment loss, classification loss. For testing phase, we only use the frame-level embedding $z$. The arrow indicates the direction of information flow.
    }
    \label{fig:architecture}
\end{figure*}

\section{Visual-Semantic Embedding Network}
We aim to learn video representations for activity retrieval, where the task is to retrieve videos of the same activity given a query video.
Let $\{(x^{(i)}, y^{(i)})\}_{i=1}^N$ be a set of $N$ activity videos, where $x$ is a video of $T$ frames describing an activity $y \in \mathcal{Y}$. Our goal is to learn an embedding function $f(\cdot) \in \mathbb{R}^{C}$ such that two different videos $x^{(i)}$ and $x^{(j)}$ of the same activity $y$ are close in the embedding space.

In large collections of activities, there usually exists an imbalance in the number of examples per activity. Following Hariharan and Girshick~\cite{hariharan2017low}, we denote activities with many examples as the \textit{base} classes and activities with few examples as the \textit{novel} classes. Formally, $\mathcal{Y}$ is then split into $\mathcal{Y}_{base}$ and $\mathcal{Y}_{novel}$, with $\mathcal{Y}_{base}\cap\mathcal{Y}_{novel} = \emptyset$.
Having an imbalanced training set causes the embedding function $f(\cdot)$ to be geared towards $\mathcal{Y}_{base}$ in the evaluation phase. As a consequence, this induces a poor retrieval performance for the under-represented classes $\mathcal{Y}_{novel}$. To alleviate this issue, we propose two alignment modules to preserve the \textit{visual} and \textit{semantic} representations of all activities.

First, we describe how to learn activity representations for all classes with a simple classification network (Section~\ref{sec:actions}). Second, we introduce two alignment modules to better handle novel classes. We propose a visual alignment module to preserve the activity representations over time (Section~\ref{sec:vis}), and a semantic alignment module to enforce activity representations to be semantically meaningful (Section~\ref{sec:sem}). Finally, we show how to train and evaluate the overall model (Section~\ref{sec:loss}).
Figure~\ref{fig:architecture} illustrates the proposed Visual-Semantic Embedding Network.

\subsection{Action representations}~\label{sec:actions}
To learn video representations of activities, we opt for a frame-level convolutional network (ConvNet) as an embedding function. Working at the frame-level rather than at the video level (\eg with 3D convolutions) offers more flexibility at the evaluation phase. In this work, frame-level representations enable us to perform localized retrieval, \eg retrieval of video clips~\cite{miech2019howto100m} or video moments~\cite{escorcia2019temporal}.

We extract the embedding representation for every frame $x_t$ and simply average them over time to obtain a video-level representation $z \in \mathbb{R}^{C}$:
\begin{equation}
\label{eq:representation}
z=\frac{1}{T} \sum_{t=1}^{T} f(x_t).
\end{equation}

The embedding representation is then further projected on a label space for classification. The probability of class $c\in\mathcal{Y}$ given an embedding representation $z$ is:
\begin{equation}
p_{\mathrm{A}}(y=c|z) = \frac{\exp(-W_c \cdot z)}{\sum_{k\in\mathcal{Y}}\exp(-W_k \cdot  z)},
\label{eq:loss1}
\end{equation}
where $W$ is the learnable parameter of the linear projection.

\subsection{Visual alignment}~\label{sec:vis}
While a standard classification embedding uses examples of all activities, the loss is in practice dominated by activities with many examples. In an effort to balance base and novel activity representations, we first focus on a visual alignment between all activities.
Let $V\in \mathbb{R}^{K\times C}$ denote a visual bank matrix consisting of features representations of dimension $C$ for every activity $y \in \mathcal{Y}$. The size of the bank then corresponds to $K=|\mathcal{Y}|$ activities. The idea of the visual bank is to obtain a single prototypical representation for every activity. Hence, all activities are treated equally, regardless of the number of examples available for training. For a new activity embedding $z$ of activity $y$, we update the visual bank $V$ through a convex combination of the current embedding representation $z$ and the corresponding entry in $V$ followed by an $\ell_2$ normalization:
\begin{equation}
\begin{split}
V_y &= \alpha \frac{z}{\|z\|_2} + (1-\alpha) V_y, \\
V_y &= V_y / \|V_y\|,
\end{split}
\label{eq:moving_average}
\end{equation}
where $\alpha$ controls the amount of update in the visual bank. The visual bank is initialized to zero when training.
%

\begin{figure}
    \centering
    \begin{overpic}[width=0.90\linewidth]{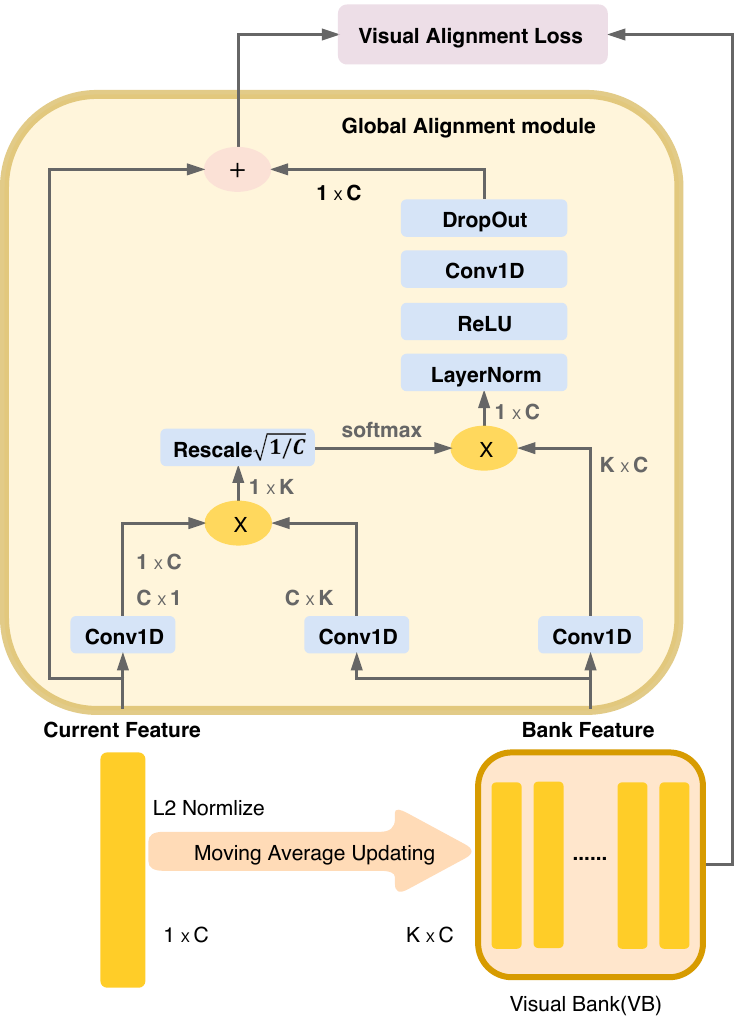}
    \end{overpic}
    
    \caption{\textbf{Visual Alignment Module}. K means class number, C denotes feature dimension. For every video feature with label y, we update the Visual Bank in the according y index. Then current feature and bank feature are fed into Global Alignment part to balance the feature from all classes. Finally a visual alignment loss is applied.   Best viewed in color. }
    \label{fig:visual_bank}
\end{figure}

Building upon such visual banks, we propose to align the representations of different activities. 
For this purpose, we rely on the attention mechanism from the non-local operator~\cite{nl}. Compared to the original non-local block, we aim to capture the relation between different prototypical representations rather than spatial~\cite{nl,silco2019} or temporal~\cite{nl,wu2019long} relations. The structure of the visual alignment module is illustrated in Figure~\ref{fig:visual_bank}.

Let $\mathrm{GA}$ denote a global alignment operator between the visual bank representation of one activity in $V$, \ie and  $\text{GA} : (\mathbb{R}^{1\times C}, \mathbb{R}^{K\times C}) \mapsto \mathbb{R}^{1\times C}$. When compared to the prototypes in the visual bank, the aligned representation $z^{\star} = \mathrm{GA}(V_c, z)$ can then be used to provide the probability of class $c$:
\begin{equation}
p_{\mathrm{V}}(y=c|z) = \frac{\exp\big(-d(z^{\star}, V_c) / \tau\big)}{\sum_{k\in\mathcal{Y}}\exp\big(-d(z^{\star}, V_k) / \tau\big)},
\label{eq:visual_alignment}
\end{equation}
where $\tau$ is the temperature of the softmax function and $d$ is the Euclidean distance.

\subsection{Semantic alignment}~\label{sec:sem}
We additionally leverage word embeddings of activity names as a prior information.
A semantic representation of an activity encapsulates relations amongst all pairs of activity classes. We use this information to additionally align activity representations towards such semantic knowledge.
We denote $\phi(y) \in \mathbb{R}^{W}$ as the word embedding of the activity $y$. Let $S\in \mathbb{R}^{K\times W}$ be the semantic bank which compiles the word embedding $\phi(y)$ of all $K$ activities.
For the semantic alignment, we simply opt for a multilayer perceptron $g(\cdot)$. Similar to the visual alignment, a probability for class $c$ can be derived after aligning the representation $z$ with the semantic activity embedding:
%
\begin{equation}
p_{\mathrm{S}}(y=c|z) = \frac{\exp\big(-d(g(z), S_c) / \tau\big)}{\sum_{k\in\mathcal{Y}}\exp\big(-d(g(z), S_c) / \tau\big)},
\label{eq:loss1}
\end{equation}

Compared to the visual bank, the semantic bank remains fixed during training. We initialize the semantic bank from an existing  word embedding (\eg word2vec~\cite{word2vec}).

\subsection{Optimization}~\label{sec:loss}
Training the overall network amounts to minimizing the cross-entropy loss function for all three components over the training set:
\begin{equation}
\mathcal{L} = -\log(p_A) -\lambda_V\log(p_V) -\lambda_S\log(p_S), 
\label{eq:loss:final}
\end{equation}
where $\lambda_V$ and $\lambda_S$ are trade-off hyper-parameters.
Once the network has been trained, we extract the video-level representations $z$ followed by an $\ell_2$ normalization for all videos in the gallery set. Similarities among videos are then measured with the Euclidean distance.

\section{Experimental Setup}

\textbf{Implementation details}
We employ ResNet-18~\cite{he2016deep} as a backbone network with weights pre-trained on ImageNet~\cite{imagenet}. Fine-tuning is done by the Adam~\cite{kingma2014adam} optimizer on one Nvidia GTX 1080TI. We set the learning rate to $1e-4$ with a weight decay of $1e-5$ for 16k iterations and reduce the learning rate to $1e-5$ after 8k iterations. We use a batch size of 16. The trade-off hyper-parameters $\lambda_{V},\lambda_{S}$ are set to 1 by cross-validation and the convex coefficient $\alpha$ of the visual bank update is set to 0.9.
Three video frames are extracted per second, resulting in an average of 32 frames per activity video. Every frame is randomly cropped and resized to $112\times 112$. We use ELMo~\cite{elmo} with 1,024 dimension as our default word embedding method.
We use PyTorch~\cite{pytorch} for implementation and the Faiss~\cite{faiss} library to measure video similarities. 

\begin{table}
    \caption{\textbf{VR-ActivityNet statistics} for video retrieval. $C_{0-100}$ are base classes with many examples, $C_{100-200}$ are novel classes with few examples.
    Some distractors with irrelevant content are constructed to simulate the real-life retrieval scenario in the testing phase.}
    \label{tab:dataset_statistics}
    \centering
    \begin{tabular}{lc|ccc}
        \toprule
        && \textbf{train} & \textbf{validation} & \cellcolor{mygray}\textbf{test} \\
        \hline
        Base&$C_{0-100}$&6095&1000&\cellcolor{mygray}3597\\ 
        \multirow{2}{*}{Novel}&$C_{100-120}$&100&200&\cellcolor{mygray}-\\
        &$C_{120-200}$&400&-&\cellcolor{mygray}2800\\
        \hline
        \multicolumn{2}{c|}{\#distractor}&-&-&\cellcolor{mygray}573 \\
        \hline
        \multicolumn{2}{c|}{Total}&6595&1200&\cellcolor{mygray}6970 \\
        \bottomrule
    \end{tabular}
\end{table}

\textbf{VR-ActivityNet} We reorganize ActivityNet1.3~\cite{caba2015activitynet} for our video retrieval and called the reorganization VR-ActivityNet. As our method aims at evaluating the performance of \textit{base} classes and \textit{novel} classes, we split the 200 activity labels into 100 base classes ($C_{0-100}$) and 100 novel classes ($C_{100-200}$). We also divide the dataset into training, validation, testing set. The validation set is trying to evaluate the balance performance between the $C_{0-100}$ and $C_{100-120}$. Similarly, the testing set is designed to evaluate the balance performance between the $C_{0-100}$ and $C_{120-200}$. Detailed activity splits are shown in the supplementary file.  

We split 10,024 untrimmed long videos from ActivityNet training set into trimmed meaningful activity segments and randomly generate a number of meaningless distractor segments. We then formulate the training and validation set of VR-ActivityNet. We utilize 4,926 untrimmed long videos from the ActivityNet validation-set to generate the trimmed segments of our testing set in VR-ActivityNet. The number of activity videos per subset in VR-ActivityNet is shown in Table~\ref{tab:dataset_statistics}. For novel classes in the training data, only 5 samples per novel class are accessible. For validation and testing data, the sample number from base and novel classes are roughly equivalent. All 6970 trimmed activity videos, except the 573 distractors, are used for retrieval. When a trimmed activity video acts as a query,  the remaining videos act as the gallery.

\textbf{Evaluation metrics.} For video retrieval, we consider the mean average precision (mAP) both on base classes and novel classes. We also compute the harmonic mean (H) between the mAP of base classes and novel classes to evaluate the balance between base and novel class performance.

\section{Results}
In the experiments, we first perform a series of ablation studies and comparisons in our proposed approach. Second, we perform further analyses to gain insight into the problem and our solution. Third, we show the ability of our model to perform retrieval of video clips and moments.

\subsection{Video retrieval experiments}~\label{sec:video}
%
\textbf{Ablation: Visual alignment.} We first investigate our visual alignment module for imbalanced activity retrieval. In Table~\ref{tab:module_ablation}, we show the results for the baseline setting which only uses a cross-entropy loss on a linear projection of the video representations, as well as the inclusion of the module. We observe an improvement of 6.2 percent point (p.p.) for base classes and 1.9 p.p. for novel classes. Hence, for both frequent and infrequent activities, the module provides a benefit.

To understand why the visual alignment module works, we investigate the discriminative abilities of activities with and without the use of the module. Ideally, prototypes should be well separated to distill discriminative information in the embedding space. To measure the scatteredness of prototypes, we calculate the $\ell_{2}$ distance of every pair of classes within base classes ($C_{0-100}$), novel classes ($C_{100-200}$), and the overall ($C_{0-200}$). The visual bank in the baseline is maintained by the Equation~\ref{eq:moving_average} without the constraint from Equation~\ref{eq:visual_alignment}. We compare the scatteredness of the visual activity proposals with and without the use of the visual alignment module. The results are shown in Table~\ref{tab:scatterness}. The scatteredness is consistently higher with our module, making all activities more unique which in turn leads to a more discriminative retrieval.
%

\begin{table}
    \caption{\textbf{Ablations} on the visual alignment module.
    }
    \label{tab:ablations}
    \centering
    \subfloat[\textbf{Visual module.} Both base and novel classes benefit from the module comapred to the baseline.\label{tab:module_ablation}]{
    \tablestyle{12pt}{1.0}
    \begin{tabular}{l|ccc}
    \toprule
    method & base & novel & \cellcolor{mygray}\textbf{H}\\
     & (mAP) & (mAP) & \cellcolor{mygray}\\
    \hline
    w/o visual&25.76&16.28&\cellcolor{mygray}19.95\\
    w/ visual&31.99&18.17&\cellcolor{mygray}23.18\\
    \bottomrule
    \end{tabular}
    }\hfill
    \subfloat[\textbf{Scatteredness.} The module works because activities are distinguished well from each other.\label{tab:scatterness}]{
    \tablestyle{12pt}{1.0}
    \begin{tabular}{l|ccc}
    \toprule
        method & base & novel & \cellcolor{mygray}overall\\
        \hline
        baseline&0.84&0.90&\cellcolor{mygray}0.87\\
        +visual &\textbf{1.19}&\textbf{1.17}&\cellcolor{mygray}\textbf{1.18}\\
        \bottomrule
    \end{tabular}
    }
\end{table}

\textbf{Ablation: Semantic alignment.}
For the semantic alignment module, we investigate its effect using four different word embedding methods. The results are shown in Table~\ref{tab:word_embedding} using  word2vec~\cite{word2vec}, ELMo~\cite{elmo},GloVe~\cite{glove}, and fasttext~\cite{fasttext}. For all word embedding methods, the  multilayer perceptron in semantic alignment module is kept the same except for the last layer.
We find that all word embeddings provide an improvement over the setting with the baseline and our visual alignment module.  For base classes, word2vec is slightly preferred, while fasttext is slightly preferred for novel classes. Overall, ELMo provides a balance between base and novel classes and we will use this word embedding for further experiments.

Having a semantic bank offers another benefit, namely an enhanced retrieval performance for different levels of the activity taxonomy. We show that this is the case by utilizing the ActivityNet taxonomy~\cite{caba2015activitynet} and evaluate the mAP for both the parent classes of the activities and the grandparent classes. The former contains 38 categories, while the latter contains 6 categories. Table~\ref{tab:taxonomy} shows that our method is able to provide improved scores for broader activity categories, highlighting that the proposed alignment results in a semantically more coherent retrieval.

\begin{table}
    \caption{\textbf{Ablations} on the semantic alignment module.
    }
    \label{tab:ablations}
    \centering
    \subfloat[\textbf{Word embeddings.} Adding a semantic prior provides an improvement, regardless of the word embedding.\label{tab:word_embedding}]{
        \tablestyle{11pt}{1.0}
    \begin{tabular}{l|ccc}
    \toprule
    method & base & novel & \cellcolor{mygray}\textbf{H}\\
     &   (mAP) & (mAP) & \cellcolor{mygray}\\
    \hline
    baseline+visual &31.99&18.17&\cellcolor{mygray}23.18\\
    +word2vec~\cite{word2vec}&33.31&18.73&\cellcolor{mygray}23.97\\
    +ELMo~\cite{elmo}&32.42&19.26&\cellcolor{mygray}24.16\\
    +GloVe~\cite{glove}&32.59&19.28&\cellcolor{mygray}24.23\\
    +Fasttext~\cite{fasttext}&32.36&19.44&\cellcolor{mygray}24.29\\
     \bottomrule
    \end{tabular}
    }\\
    \subfloat[\textbf{Retrieval result in various activity taxonomy hierarchy}. Level-1 contains 6 super classes, level-2 contains 38 super classes. The mAP is evaluated on the overall classes.\label{tab:taxonomy}]{
    \tablestyle{11pt}{1.0}
    \begin{tabular}{l|cc}
    \toprule
      method & level-1(6 -cls) & level-2(38-cls)\\
       & (mAP) & (mAP)\\
    \hline
    baseline+visual&22.41&20.35\\
    \rowcolor{mygray}
    +semantic&\textbf{23.14}&\textbf{21.76} \\
        \bottomrule
    \end{tabular}
    }
\end{table}

\textbf{Comparison with other methods.}
Using our two modules, we perform a comparative evaluation to three baseline retrieval approaches. The first baseline serves as a starting point. We use the network used in this work but only pre-trained on ImageNet~\cite{imagenet} to obtain video representations by averaging their frames. Query and candidate videos are then matched using the Euclidean distance. As result in Table~\ref{tab:comparison_baselinne} shows, the low scores indicate the difficulty of the task. Interestingly, the off-the-shelf baseline doesn't suffer from an imbalance performance between the base classes and novel classes. This confirms the fact that when fine-tuning, representations of videos then tend to be more discriminative towards the base classes as they are more frequent.

Table~\ref{tab:comparison_baselinne} also shows the consequence of imbalanced fine-tuning for two accepted approaches in retrieval, namely the triplet loss~\cite{triplenet} and the margin loss\cite{marginloss} optimized on top of the same video representations as for our approach. Both approaches obtain a boost in base mAP and a smaller improvement in novel mAP. Both the sampling-based loss baselines and our baseline setup do not explicitly cater to novel classes, resulting in similar scores for the harmonic mean. Our proposed approach performs favorably compared to all baselines, both for base and for novel classes. Our harmonic mAP is respectively 13.4, 4.5, 3.4, and 4.2 percent point higher than the baselines. We conclude that our formulation is preferred for activity retrieval regardless of whether they have many or few examples to train on.
%

\begin{table}
    \caption{\textbf{Comparison with other methods.} Our approach is preferred over both internal and external baselines, since our modules explicitly give equal importance to base and novel classes.}
    \label{tab:comparison_baselinne}
    \centering
    \begin{tabular}{l|ccc}
        \toprule
        & base & novel& \cellcolor{mygray}\textbf{H}\\
         &   (mAP) & (mAP) & \cellcolor{mygray}\\
        \hline
        ImageNet~\cite{imagenet}&9.18&13.02&\cellcolor{mygray}10.76\\
        Triple loss~\cite{triplenet}&24.47&16.48&\cellcolor{mygray}19.70\\
        Margin loss~\cite{marginloss}&25.84&17.36&\cellcolor{mygray}20.76\\
        \hline
        baseline &25.76&16.28&\cellcolor{mygray}19.95\\
        w/ our modules&\textbf{32.42}&\textbf{19.26}&\cellcolor{mygray}\textbf{24.16}\\
        \bottomrule
    \end{tabular}
\end{table}

\subsection{Video retrieval analyses}~\label{sec:video}
We perform three analyses to gain insight into the imbalanced activity retrieval problem and into our approach.

\begin{figure}
     \centering
     \includegraphics[width=.8\columnwidth]{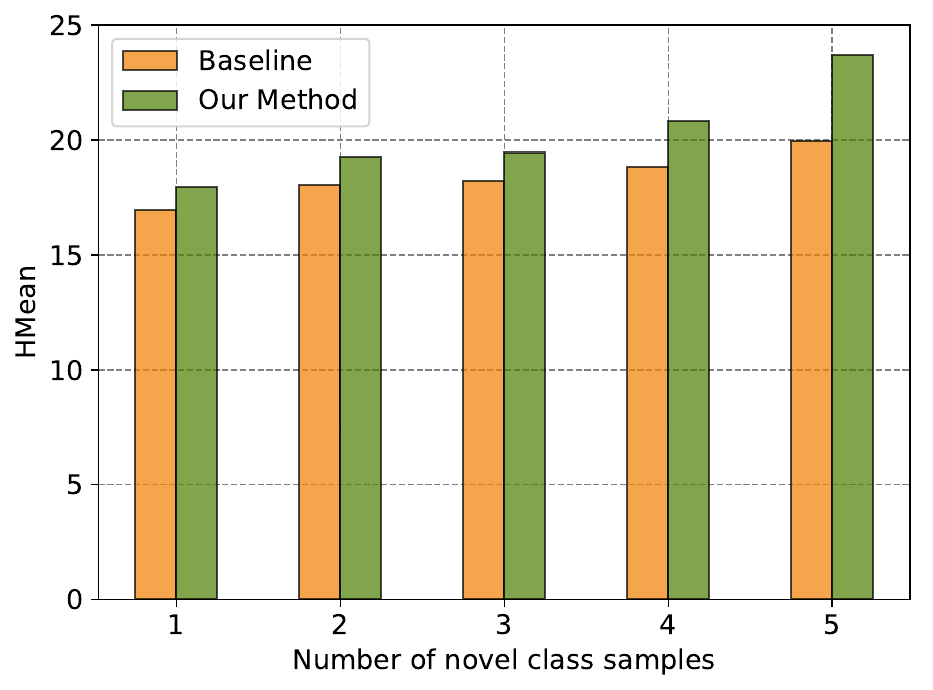}
     \caption{\textbf{Effect of the number of samples per novel class.} With our modules, the improvement gap increases with more examples per novel activity. Our balanced optimization is not only beneficial for rare activities, but also for more frequent ones.}
     \label{fig:gradual_novel_instances}
 \end{figure}

\begin{figure}
     \centering
     \includegraphics[width=.8\columnwidth]{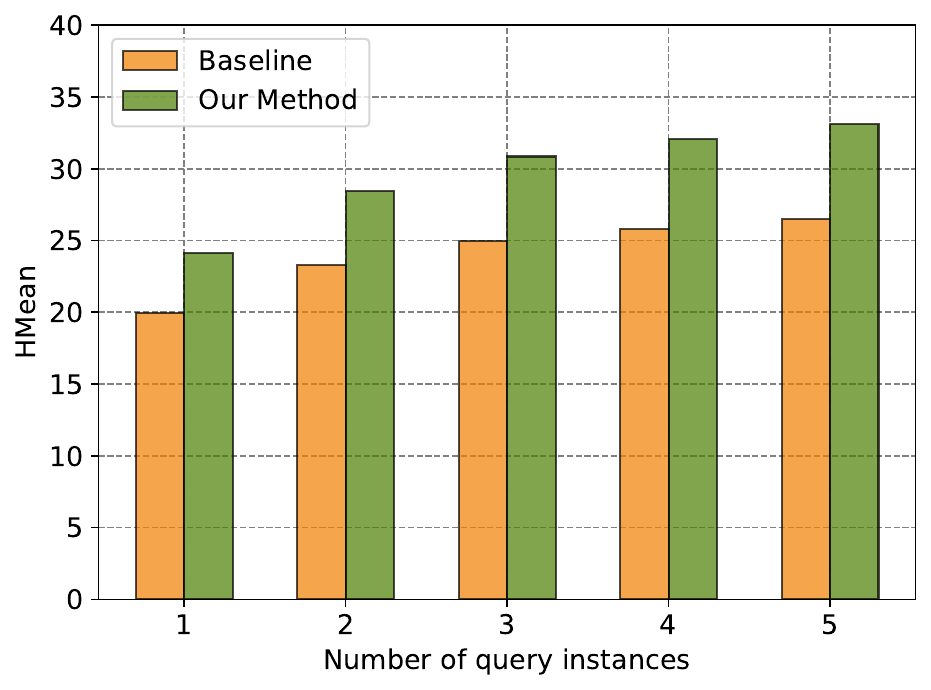}
     \caption{\textbf{Effect of the number of query videos per retrieval.} We gradually increase the number of queries from 1 to 5. Our approach is effective for both the standard and multi-shot scenario.}
     \label{fig:gradual_query_instances}
 \end{figure}

\textbf{Increasing the number of samples.} First, we study the effect of the number of samples per novel class during training in Figure~\ref{fig:gradual_novel_instances}. We find that even when one novel class sample is provided, our method can distill knowledge from limited provided supervision. As the number of examples for novel classes increases, the gap with the baseline also increases, highlighting that our balanced formulation also helps for many examples.
We also study the effect of the number of activity videos per query during the testing phase. When using more than one query video, we average the features of all queries before retrieval. Figure~\ref{fig:gradual_query_instances} shows that a consistent gain can be collected when increasing the number of query videos, which shows our method benefits from having multiple videos as a query.

\textbf{Robustness to data splits.} As generalization over new splits is not necessarily achieved in the presence of novel classes~\cite{chen19closerfewshot,shchur2018pitfalls}, we evaluate our proposed model on two other data splits. Table ~\ref{tab:imbalance_case} shows the results for the two following settings: (B120, N80)  and (B80, N100), where B denotes the number of base classes and N the number of novel classes. The consistent improvements across these new splits indicate that our approach is not tuned to specific splits and can work whether we have many or few infrequent activities.

\begin{table}
    \caption{\textbf{More dataset splitting case.} We evaluate on two diffent class label splittings: (B120, N80) and (B80, N120). Note that the original dataset splitting is (B100, N100). Our method is consistent over three different dataset splits.}
    \label{tab:imbalance_case}
    \centering
    \begin{tabular}{l|ccc}
        \toprule
        & base & novel & \cellcolor{mygray}\textbf{H}\\
        &(mAP)&(mAP)&\cellcolor{mygray}\\
        \hline
        \textbf{(B120, N80)}&&& \cellcolor{mygray}\\
        baseline&22.72&13.41&\cellcolor{mygray}16.86\\
        baseline+visual&28.15&14.72&\cellcolor{mygray}19.33\\
        baseline+visual+semantic &\textbf{29.38}&\textbf{14.91}&\cellcolor{mygray}\textbf{19.78}\\
        \hline
        \textbf{(B80, N120)}&&& \cellcolor{mygray}\\
        baseline&27.39&14.74&\cellcolor{mygray}19.16\\
        baseline+visual&33.14&17.14&\cellcolor{mygray}22.59\\
        baseline+visual+semantic &\textbf{33.85}&\textbf{17.63}&\cellcolor{mygray}\textbf{23.18}\\
        \bottomrule
    \end{tabular}
\end{table}

\begin{figure}
    \centering
    \includegraphics[width=0.9\columnwidth]{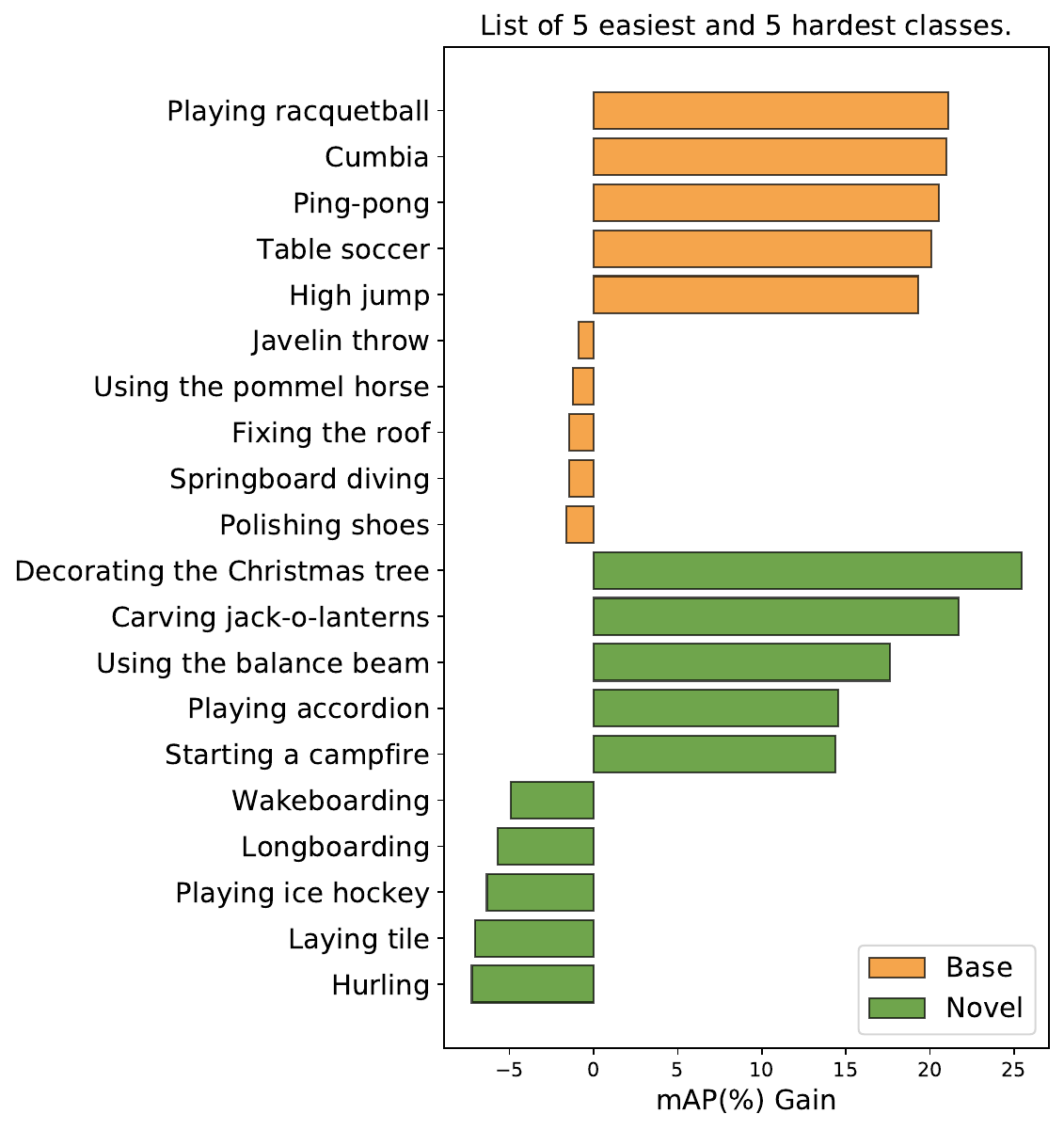}
    \caption{\textbf{The gain and loss analysis.} We pick out the 5 easiest and 5 hardest classes from base and novel classes respectively. The x-axis is the relative gain in percentage.}
    \label{fig:gain_loss_perclass}
\end{figure}

\textbf{Qualitative analysis.} Intuitively, not all activities benefit from the inclusion of our visual and semantic alignments for balancing activities. In Figure~\ref{fig:gain_loss_perclass}, we show which classes benefit and suffer the most after applying the two alignment modules. We select the 5 easiest and 5 hardest classes from the base and novel classes respectively. For the novel classes, gains are important for fine-grained activities with a salient object, such as \textit{decorating the Christmas tree} or \textit{carving jack-o-lanterns}. For the base classes, gains are important for sports activities. Indeed, a fine-grained understanding is required to differentiate among these activities and having both alignment modules helps to separate them. We observe that our approach suffers for multiple sports activities with few examples, showing the direct downside of the boost for sports activities with many examples, as they will become a more likely retrieval candidate given a query.

\begin{figure*}
    \centering
    \includegraphics[width=0.94\textwidth]{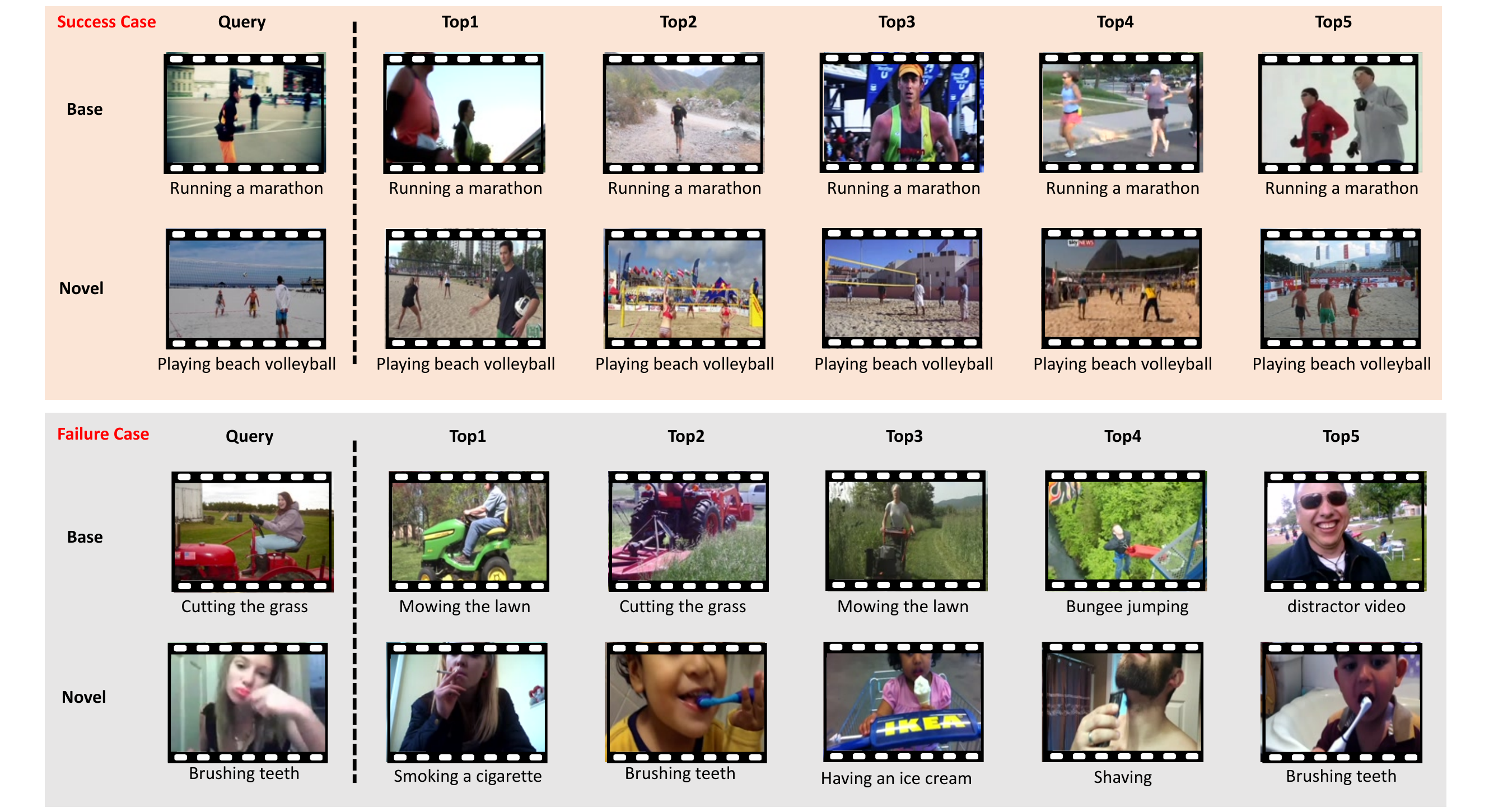}
    \caption{\textbf{Retrieval visualization of our method.} The upper part shows two successful cases while the bottom two failure cases. We evaluate with either a query from a base or novel class. The two successful cases demonstrate our method can tackle the different background distractors and extract the essential information. The failing \textit{cutting the grass} is distracted by green grass in distractor video, green tree in \textit{bungee junmping}, context information in \textit{mowing the lawn}. The failure case in \textit{Brushing teeth} contains a similar context information, while still fails. The object recognition of \textit{cigarette}, \textit{ice cream}, or \textit{shaves} would be helpful for the retrieval task. }
    \label{fig:visualization}
\end{figure*}

Figure~\ref{fig:visualization} also presents successful and failure cases. For the two success cases, our method can tackle various background distractors to extract  essential video information. For the failure case of \textit{cutting the grass}, our method is distracted by \eg the tree in the \textit{bungee jumping} example and by the highly-similar activity \textit{mowing the lawn}. For the failure case of \textit{brushing teeth}, the context information to other activities is very similar, while small key objects such as \textit{cigarette}, \textit{ice cream}, \textit{shaver} are ignored by our method. Having object information could further boost the retrieval performance.

\subsection{Clip and moment retrieval}~\label{sec:clip+moment}
Beyond retrieving videos, our approach is also suitable for retrieving video clips and video moments, both of which have recently gained traction. In a retrieval context, video clips denote local video segments of a fixed length~\cite{miech2019howto100m}, while video moments denote localized segments marking the duration of the activity in a whole video~\cite{anne2017localizing}.



\begin{table}
    \caption{\textbf{Clip retrieval evaluation} for clips of 4, 6, and 8 seconds. We find favourable results for all clip lengths, especially when videos are longer.}
    \label{tab:clip_duration}
    \centering
    \begin{tabular}{l|ccc}
        \toprule
        clip-duration & base & novel & \textbf{H}\\
         &  (mAP) & (mAP) & \\
        \hline
        \rowcolor{mygray}\textbf{4 seconds}&&&\\
        Margin loss~\cite{marginloss}&14.06&10.10&11.76\\
        \hline
        baseline&13.38&10.33&11.66\\
        our method&\textbf{17.62}&\textbf{12.85}&\textbf{14.86}\\
        \hline
        \hline
        \rowcolor{mygray}\textbf{6 seconds}&&&\\
        Margin loss~\cite{marginloss}&14.80&10.94&12.58\\
        \hline
        baseline&13.62&10.76&12.03\\
        Our method &\textbf{18.19}&\textbf{13.36}&\textbf{15.40}\\
        \hline
        \hline
        \rowcolor{mygray}\textbf{8 seconds}&&&\\
        Margin loss~\cite{marginloss}&15.23&11.32&12.99\\
        \hline
        baseline&13.96&10.99&12.30\\
        our method&\textbf{18.65}&\textbf{13.75}&\textbf{15.83}\\
        \bottomrule
    \end{tabular}
\end{table}

\begin{figure}
    \centering
    \includegraphics[width=.7\linewidth]{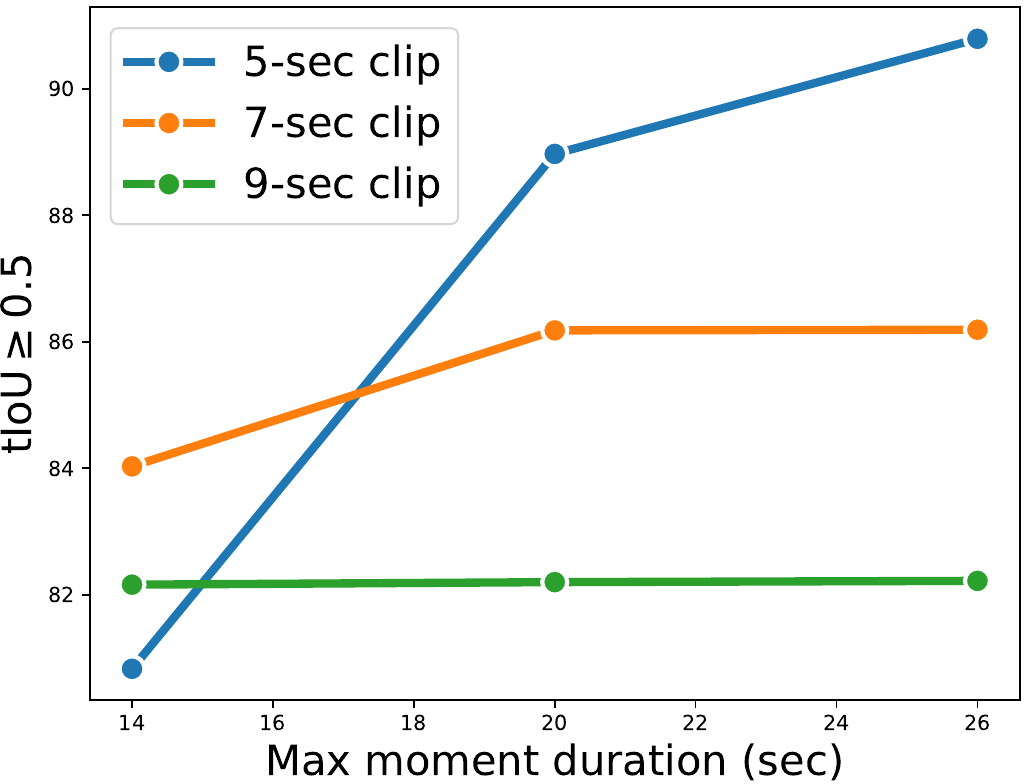}
    \caption{\textbf{VR-ActivityNet statistics about the max moment length (M) and clip length (N).} We exhaustively list all possible moment temporal proposals in all videos of our dataset. A hit occurs when the tIoU between a proposal with class $\textit{C}$ and one ground truth proposal with class $\textit{C}$ is larger than 0.5. The best combination is N=5, M=26, which is our default setting in moment retrieval task.}
    \label{fig:activitynet_hack}
\end{figure}

\begin{table}
    \caption{\textbf{Performance on Moment Retrieval.} Our method can perform favorably compared with our baseline.}
    \label{tab:moment_retrieval}
    \centering
    \begin{tabular}{l|ccc}
        \toprule
        &base&novel&\cellcolor{mygray}\textbf{H}\\
         &  (mAP) & (mAP) & \cellcolor{mygray}\\
        \hline
        Margin loss~\cite{marginloss}&7.06&5.66&\cellcolor{mygray}6.28\\
        \hline
        baseline&8.44&7.03&\cellcolor{mygray}7.67\\
        Our method&\textbf{9.14}&\textbf{7.15}&\cellcolor{mygray}\textbf{8.02}\\
        \bottomrule
    \end{tabular}
\end{table}

\textbf{Clip retrieval.} For clip retrieval, we set the fixed duration to 4, 6, and 8 seconds. All videos are split into fixed-length clips, where a clip is positive if its temporal during lies within the boundaries of the activity. The retrieval is performed over all individual clips.

We show clip retrieval results on the same dataset as video retrieval in Table~\ref{tab:clip_duration}. We use the most effective video retrieval baseline, the margin loss~\cite{marginloss}, as a baseline here. We find that regardless of the clip duration, our approach is preferred. As video clips become longer, the performance gap slightly increases for base and novel activities. Overall, we conclude that a generalization to video clips for retrieval is viable for our approach.


\textbf{Moment retrieval.}
Lastly, we investigate localized video moment retrieval with our approach. We obtain temporal proposals by starting from video clips and performing a sliding window over all sets of consecutive clips exhaustively. We observe that such a proposal setup readily obtains proposals with high recall, as shown in Figure~\ref{fig:activitynet_hack}. For retrieval, we score each proposal in each video and rank them by similarity score. Table~\ref{tab:moment_retrieval} shows that our method can also generalize to video moment retrieval with imbalanced activities. The improvements are more marginal compared to video and clip retrieval, since moment retrieval entails a more difficult task due to the additional temporal localization.


\section{Conclusion}
In this work we propose a new task about video retrieval by activity in the wild, and emphasize the importance of dealing with imbalanced data when retrieving activities from a video query. We introduce an embedding network that learns to balance frequent base activities and infrequent novel activities. The network contains two novel modules. A visual alignment module matches input videos with visual prototype representations of activities. A semantic alignment module on the other hand matches videos with word embedding representations of activities. Visual and semantic activity representations are of the same length, regardless of the number of examples each activity has. As a result, we arrive at an activity retrieval that better balances both types of activities. We show this result empirically by proposing a new imbalanced activity retrieval dataset with a revised  data splits. Experiments highlight the effectiveness of our approach, as well as a series of ablations and analyses to gain insight into the problem. Lastly, we show how our approach generalizes to video clip and moment retrieval from video queries in imbalanced settings.

{\small
\bibliographystyle{ieee_fullname}
\bibliography{6-egbib}

\begin{thebibliography}{10}\itemsep=-1pt

\bibitem{bart2005cross}
Evgeniy Bart and Shimon Ullman.
\newblock Cross-generalization: Learning novel classes from a single example by
  feature replacement.
\newblock In {\em CVPR}, 2005.

\bibitem{BuloNK17}
Samuel~Rota Bul{\`{o}}, Gerhard Neuhold, and Peter Kontschieder.
\newblock Loss max-pooling for semantic image segmentation.
\newblock {\em CoRR}, 2017.

\bibitem{caba2015activitynet}
Fabian Caba~Heilbron, Victor Escorcia, Bernard Ghanem, and Juan Carlos~Niebles.
\newblock Activitynet: A large-scale video benchmark for human activity
  understanding.
\newblock In {\em CVPR}, 2015.

\bibitem{chen19closerfewshot}
Wei-Yu Chen, Yen-Cheng Liu, Zsolt Kira, Yu-Chiang Wang, and Jia-Bin Huang.
\newblock A closer look at few-shot classification.
\newblock In {\em ICCV}, 2019.

\bibitem{ciptadi2014movement}
Arridhana Ciptadi, Matthew~S Goodwin, and James~M Rehg.
\newblock Movement pattern histogram for action recognition and retrieval.
\newblock In {\em ECCV}, 2014.

\bibitem{imagenet}
Jia Deng, Wei Dong, Richard Socher, Li-Jia Li, Kai Li, and Li Fei-Fei.
\newblock {ImageNet}: A large-scale hierarchical image database.
\newblock In {\em CVPR}, 2009.

\bibitem{douze2016circulant}
Matthijs Douze, J{\'e}r{\^o}me Revaud, Jakob Verbeek, Herv{\'e} J{\'e}gou, and
  Cordelia Schmid.
\newblock Circulant temporal encoding for video retrieval and temporal
  alignment.
\newblock {\em IJCV}, 2016.

\bibitem{escorcia2019temporal}
Victor Escorcia, Mattia Soldan, Josef Sivic, Bernard Ghanem, and Bryan Russell.
\newblock Temporal localization of moments in video collections with natural
  language.
\newblock {\em arXiv}, 2019.

\bibitem{maml}
Chelsea Finn, Pieter Abbeel, and Sergey Levine.
\newblock Model-agnostic meta-learning for fast adaptation of deep networks.
\newblock {\em ICML}, 2017.

\bibitem{gao2017tall}
Jiyang Gao, Chen Sun, Zhenheng Yang, and Ram Nevatia.
\newblock Tall: Temporal activity localization via language query.
\newblock In {\em ICCV}, 2017.

\bibitem{hariharan2017low}
Bharath Hariharan and Ross Girshick.
\newblock Low-shot visual recognition by shrinking and hallucinating features.
\newblock In {\em ICCV}, 2017.

\bibitem{he2016deep}
Kaiming He, Xiangyu Zhang, Shaoqing Ren, and Jian Sun.
\newblock Deep residual learning for image recognition.
\newblock In {\em CVPR}, 2016.

\bibitem{anne2017localizing}
Lisa~Anne Hendricks, Oliver Wang, Eli Shechtman, Josef Sivic, Trevor Darrell,
  and Bryan Russell.
\newblock Localizing moments in video with natural language.
\newblock In {\em ICCV}, 2017.

\bibitem{hendricks2018localizing}
Lisa~Anne Hendricks, Oliver Wang, Eli Shechtman, Josef Sivic, Trevor Darrell,
  and Bryan Russell.
\newblock Localizing moments in video with temporal language.
\newblock {\em arXiv}, 2018.

\bibitem{triplenet}
Elad Hoffer and Nir Ailon.
\newblock Deep metric learning using triplet network.
\newblock In {\em International Workshop on Similarity-Based Pattern
  Recognition}, 2015.

\bibitem{silco2019}
Tao Hu, Pascal Mettes, Jia-Hong Huang, and Cees~GM Snoek.
\newblock Silco: Show a few images, localize the common object.
\newblock In {\em ICCV}, 2019.

\bibitem{faiss}
Jeff Johnson, Matthijs Douze, and Herv{\'e} J{\'e}gou.
\newblock Billion-scale similarity search with gpus.
\newblock {\em arXiv}, 2017.

\bibitem{fasttext}
Armand Joulin, Edouard Grave, Piotr Bojanowski, and Tomas Mikolov.
\newblock Bag of tricks for efficient text classification.
\newblock {\em arXiv}, 2016.

\bibitem{kang2003query}
In-Ho Kang and GilChang Kim.
\newblock Query type classification for web document retrieval.
\newblock In {\em ACM Conference on Research and Development in Information
  Retrieval}, 2003.

\bibitem{kervadec19a}
Hoel Kervadec, Jihene Bouchtiba, Christian Desrosiers, Eric Granger, Jose Dolz,
  and Ismail {Ben Ayed}.
\newblock Boundary loss for highly unbalanced segmentation.
\newblock In {\em Conference on Medical Imaging with Deep Learning}, 2019.

\bibitem{kingma2014adam}
Diederik~P Kingma and Jimmy Ba.
\newblock Adam: A method for stochastic optimization.
\newblock {\em ICLR}, 2016.

\bibitem{liu2019large}
Ziwei Liu, Zhongqi Miao, Xiaohang Zhan, Jiayun Wang, Boqing Gong, and Stella~X
  Yu.
\newblock Large-scale long-tailed recognition in an open world.
\newblock In {\em CVPR}, 2019.

\bibitem{luo2019few}
Tiange Luo, Aoxue Li, Tao Xiang, Weiran Huang, and Liwei Wang.
\newblock Few-shot learning with global class representations.
\newblock {\em arXiv}, 2019.

\bibitem{mettes2019hyperspherical}
Pascal Mettes, Elise van~der Pol, and Cees G.~M. Snoek.
\newblock Hyperspherical prototype networks.
\newblock {\em arXiv}, 2019.

\bibitem{miech2019howto100m}
Antoine Miech, Dimitri Zhukov, Jean-Baptiste Alayrac, Makarand Tapaswi, Ivan
  Laptev, and Josef Sivic.
\newblock Howto100m: Learning a text-video embedding by watching hundred
  million narrated video clips.
\newblock In {\em ICCV}, 2019.

\bibitem{word2vec}
Tomas Mikolov, Ilya Sutskever, Kai Chen, Greg~S Corrado, and Jeff Dean.
\newblock Distributed representations of words and phrases and their
  compositionality.
\newblock In {\em NeurIPS}, 2013.

\bibitem{mithun2019weakly}
Niluthpol~Chowdhury Mithun, Sujoy Paul, and Amit~K Roy-Chowdhury.
\newblock Weakly supervised video moment retrieval from text queries.
\newblock In {\em CVPR}, 2019.

\bibitem{oksuz2019imbalance}
Kemal Oksuz, Baris~Can Cam, Sinan Kalkan, and Emre Akbas.
\newblock Imbalance problems in object detection: A review.
\newblock {\em arXiv}, 2019.

\bibitem{otani2016learning}
Mayu Otani, Yuta Nakashima, Esa Rahtu, Janne Heikkil{\"a}, and Naokazu Yokoya.
\newblock Learning joint representations of videos and sentences with web image
  search.
\newblock In {\em ECCV}, 2016.

\bibitem{pytorch}
Adam Paszke, Sam Gross, Soumith Chintala, Gregory Chanan, Edward Yang, Zachary
  DeVito, Zeming Lin, Alban Desmaison, Luca Antiga, and Adam Lerer.
\newblock Automatic differentiation in pytorch.
\newblock In {\em NeurIPS}, 2017.

\bibitem{glove}
Jeffrey Pennington, Richard Socher, and Christopher Manning.
\newblock Glove: Global vectors for word representation.
\newblock In {\em EMNLP}, 2014.

\bibitem{elmo}
Matthew~E Peters, Mark Neumann, Mohit Iyyer, Matt Gardner, Christopher Clark,
  Kenton Lee, and Luke Zettlemoyer.
\newblock Deep contextualized word representations.
\newblock {\em arXiv}, 2018.

\bibitem{qin2017fast}
Jie Qin, Li Liu, Mengyang Yu, Yunhong Wang, and Ling Shao.
\newblock Fast action retrieval from videos via feature disaggregation.
\newblock {\em Computer Vision and Image Understanding}, 2017.

\bibitem{rusu2018meta}
Andrei~A Rusu, Dushyant Rao, Jakub Sygnowski, Oriol Vinyals, Razvan Pascanu,
  Simon Osindero, and Raia Hadsell.
\newblock Meta-learning with latent embedding optimization.
\newblock {\em arXiv}, 2018.

\bibitem{serra2018overcoming}
Joan Serr{\`a}, D{\'\i}dac Sur{\'\i}s, Marius Miron, and Alexandros
  Karatzoglou.
\newblock Overcoming catastrophic forgetting with hard attention to the task.
\newblock {\em arXiv}, 2018.

\bibitem{shchur2018pitfalls}
Oleksandr Shchur, Maximilian Mumme, Aleksandar Bojchevski, and Stephan
  G{\"u}nnemann.
\newblock Pitfalls of graph neural network evaluation.
\newblock {\em arXiv}, 2018.

\bibitem{prototypical}
Jake Snell, Kevin Swersky, and Richard Zemel.
\newblock Prototypical networks for few-shot learning.
\newblock In {\em NeurIPS}, 2017.

\bibitem{snoek2009concept}
Cees~GM Snoek, Marcel Worring, et~al.
\newblock Concept-based video retrieval.
\newblock {\em Foundations and Trends in Information Retrieval}, 2009.

\bibitem{song2018self}
Jingkuan Song, Hanwang Zhang, Xiangpeng Li, Lianli Gao, Meng Wang, and Richang
  Hong.
\newblock Self-supervised video hashing with hierarchical binary auto-encoder.
\newblock {\em IEEE Transactions on Image Processing}, 2018.

\bibitem{torabi2016learning}
Atousa Torabi, Niket Tandon, and Leonid Sigal.
\newblock Learning language-visual embedding for movie understanding with
  natural-language.
\newblock {\em arXiv}, 2016.

\bibitem{nl}
Xiaolong Wang, Ross Girshick, Abhinav Gupta, and Kaiming He.
\newblock Non-local neural networks.
\newblock In {\em CVPR}, 2018.

\bibitem{wu2019long}
Chao-Yuan Wu, Christoph Feichtenhofer, Haoqi Fan, Kaiming He, Philipp
  Krahenbuhl, and Ross Girshick.
\newblock Long-term feature banks for detailed video understanding.
\newblock In {\em CVPR}, 2019.

\bibitem{marginloss}
Chao-Yuan Wu, R Manmatha, Alexander~J Smola, and Philipp Krahenbuhl.
\newblock Sampling matters in deep embedding learning.
\newblock In {\em ICCV}, 2017.

\bibitem{xian2017zero}
Yongqin Xian, Bernt Schiele, and Zeynep Akata.
\newblock Zero-shot learning-the good, the bad and the ugly.
\newblock In {\em CVPR}, 2017.

\bibitem{xu2019multilevel}
Huijuan Xu, Kun He, Bryan~A Plummer, Leonid Sigal, Stan Sclaroff, and Kate
  Saenko.
\newblock Multilevel language and vision integration for text-to-clip
  retrieval.
\newblock In {\em AAAI}, 2019.

\bibitem{zhu2018compound}
Linchao Zhu and Yi Yang.
\newblock Compound memory networks for few-shot video classification.
\newblock In {\em ECCV}, 2018.

\bibitem{zhu2014capturing}
Xiangxin Zhu, Dragomir Anguelov, and Deva Ramanan.
\newblock Capturing long-tail distributions of object subcategories.
\newblock In {\em CVPR}, 2014.

\end{thebibliography}
}


\clearpage
\onecolumn
\section{Supplementary File}
\tableofcontents


\textbf{A. Extra experimental results}

\begin{figure}[t]
	\centering
	\includegraphics[width=1.0\columnwidth]{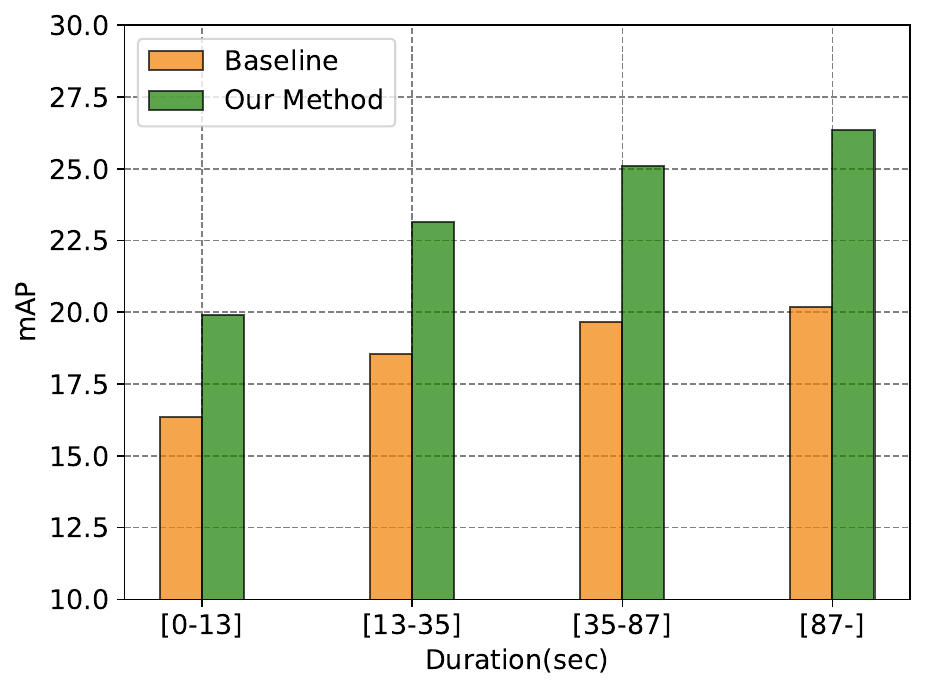}
	\caption{\textbf{Retrieval performance with respect to the query video duration (sec).} Our method is able to collect consistent gain as duration increases, which demonstrates a better feature fusion efficacy can be achieved by our method.}
	\label{fig:duration_ablation}
\end{figure}

\textbf{A.1 Duration analysis.} We analyze the performance with respect to the duration of the query videos in Figure~\ref{fig:duration_ablation}. We can observe that the longer the query video is, the more performance gain our method can collect, which indicates the efficacy of our method.

\begin{figure}[t]
	\centering
	\includegraphics[width=1.0\columnwidth]{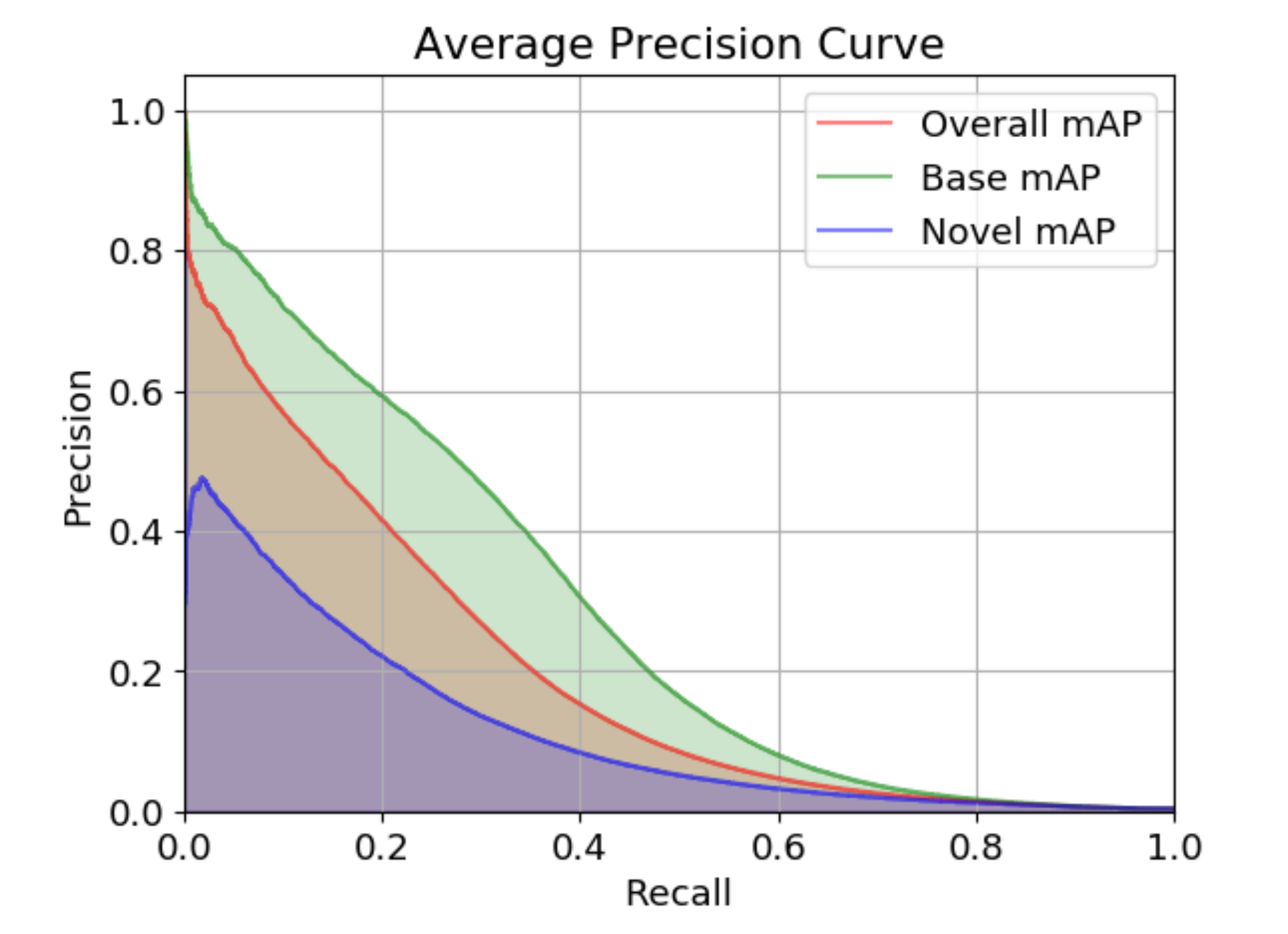}
	\caption{\textbf{mAP curve of our best result,} which shows the base, novel and overall classes.}
	\label{fig:map_all}
\end{figure}

\textbf{A.2 mAP curve.}  mAP is an important metric to evaluate the balance between the precision and recall of retrieval task. All retrieval results per query are considered in mAP. We depict the mAP in Figure~\ref{fig:map_all}. The performance gap between base and novel classes is still relatively large, more improvement such as object recognition ability can be further embedded to boost the performance.

\begin{figure*}
	\centering
	\includegraphics[width=1.0\textwidth]{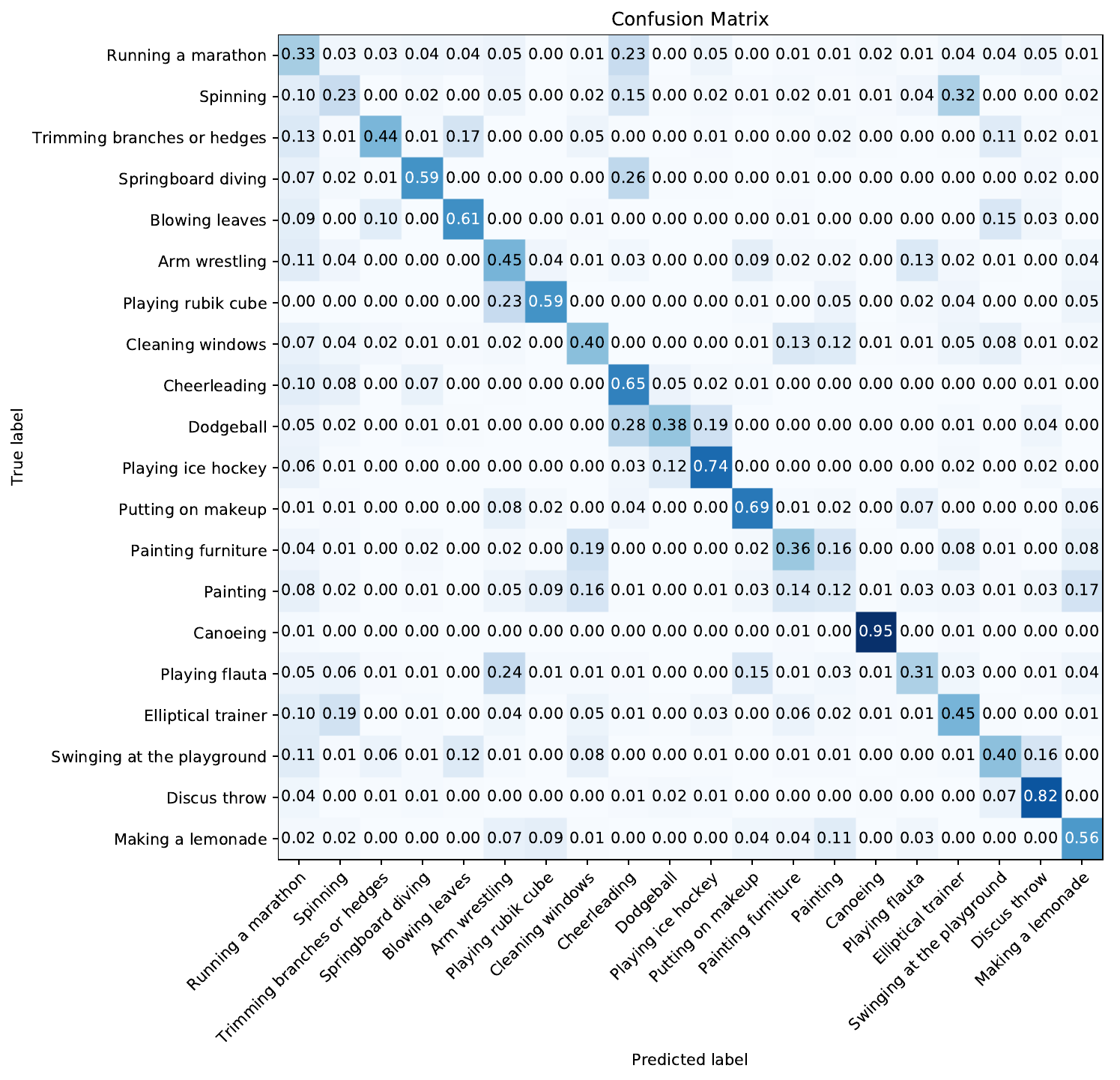}
	\caption{\textbf{Confusion matrix} demonstrate that our method can deal with most of the cases well. Vertical axis means ground truth, horizontal axis denotes prediction.  The 20 classes are random selected from the 200 classes. The top-100 will seen as a hit in the generation of confusion matrix.}
	\label{fig:confusion_matrix}
\end{figure*}

\textbf{A.3 Confusion matrix.} As the original number of activity class labels is 200, we randomly select 20 classes to illustrate the confusion matrix in Figure~\ref{fig:confusion_matrix}. The top-100  retrieval will be all seen as a hit in the generation of confusion matrix. We observe that \textit{canoeing} performs best because the context information is not so complex, while \textit{painting} presents the worst result among those classes. This probably comes from the context information in \textit{painting}, which is more  variable. Also, we can obverse that \textit{arm wrestling} is seriously mistaken as \textit{playing flauta} as both activities show arms. These observations also align with the phenomenon in the visualization results. The ability of object recognition could further boost the performance of this task. 

\begin{figure}
	\centering
	\includegraphics[width=0.9\columnwidth]{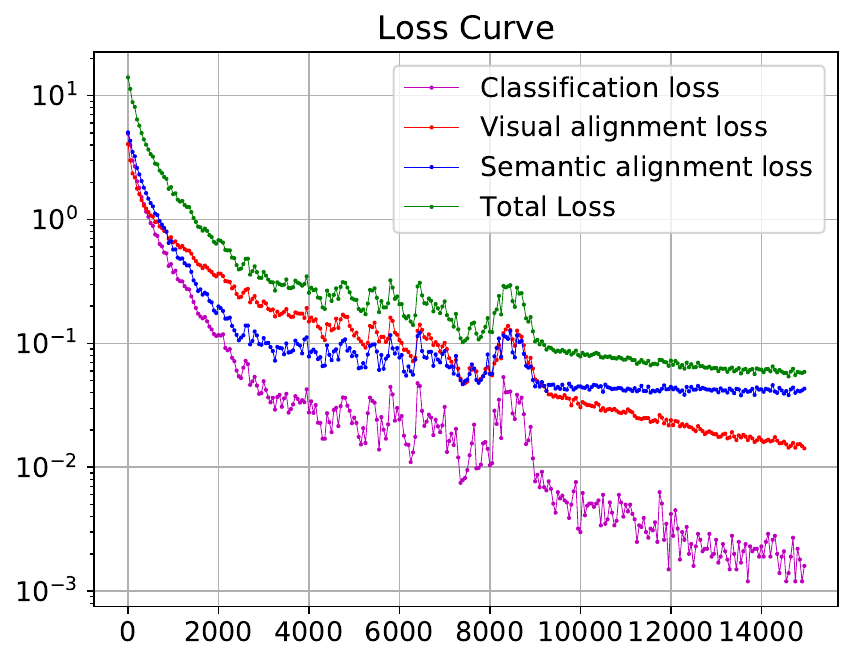}
	\caption{\textbf{Loss Curve.} We show four items: classification loss, visual alignment loss, semantic alignment loss and total loss. Log scale in Y-axis.}
	\label{fig:loss}
\end{figure}

\textbf{A.4 Loss curve.} We show our loss decreasing curve in Figure~\ref{fig:loss}. Three different kinds of loss items are depicted. A globally stable decreasing trend can be observed.

\textbf{A.5 Dataset label splits.} We summarize the dataset label split in Table~\ref{tab:dataset-label}. Those labels are randomly splited into three subsets.

\textbf{B. Implementation details}

\textbf{B.1 Evaluation Metric Calculation.} We use mAP as the main metric, for a better balance of performance between base and novel classes, similar to ~\cite{xian2017zero}, we consider the harmonic mean between the base and novel mAP  as our evaluation metrics. The calculation of mAP of base classes and novel classes is:

\begin{equation}
    mAP = \frac{1}{M}\sum_{i=1}^{K}\sum_{j=1}^{Q_{c}} AP_{i,j}
\end{equation}

where K is the total class number, $Q_{k}$ is the query number from class k, $AP_{k,j}$ denotes the AP of class from k, query from j. M is the total query video number from base or novel classes.

\textbf{B.2 Baseline details.} \textbf{Triplet loss.} We randomly sample 2 classes to formulate the anchor, positive, negative class. Based on these classes, we further randomly sample three videos to construct a triplet for training. \textbf{Margin loss~\cite{marginloss}.} The margin parameter is set to 0.2.

\textbf{B.3 FC Layer in Semantic Alignment module.} Let FC(M) denote a  fully connected layer with M units. ReLU denotes the activation function and S means the dimension of the specific word embedding method.  The FC Layer in Semantic Alignment Module can be formulated as: FC(512)-ReLU-FC(640)-ReLU-FC(768)-ReLU-FC(896)-ReLU-FC(S).

\textbf{B.4 ActivityNet hierarchy label preprocessing.} Our hierarchy label is slightly different from the official annotation of ActivityNet~\cite{caba2015activitynet}, we tidy up the hierarchy structure based on the official version to formulate a 6 grandfather activity categories, 38 father activity categories, and 200 activity classes. Our hierarchy structure will be released public upon acceptance.

\begin{table*}
	\caption{\textbf{Dataset labels.} The validation set is trying to evaluate the balance performance between the $C_{0-100}$ and $C_{100-120}$, similarly, the testing set is designed to evaluate the balance performance between the $C_{0-100}$ and $C_{120-200}$. }
	\label{tab:dataset-label}
	\centering
	\begin{tabularx}{\textwidth}{l|X}
		\toprule
		subset & labels \\
		\hline
		\textbf{Base: $C_{0-100}$}& Cricket, Cleaning windows, Cutting the grass, Roof shingle removal, Cheerleading, Skiing, Using parallel bars, Putting in contact lenses, Doing step aerobics, Bathing dog, Springboard diving, Camel ride, Horseback riding, Installing carpet, Washing dishes, Grooming dog, Getting a piercing, Rafting, Using the monkey bar, Rock-paper-scissors, Ping-pong, Washing hands, Arm wrestling, Putting on shoes, Snatch, Doing fencing, Chopping wood, Fixing the roof, Doing kickboxing, River tubing, Belly dance, Spinning, Waxing skis, Layup drill in basketball, Ice fishing, Zumba, Slacklining, Sumo, Kayaking, Playing congas, Doing nails, Playing harmonica, High jump, Playing racquetball, Table soccer, Preparing salad, Running a marathon, Doing motocross, Smoking a cigarette, Hitting a pinata, Polishing forniture, Tennis serve with ball bouncing, Hula hoop, Disc dog, Hopscotch, Smoking hookah, Bullfighting, Hammer throw, Javelin throw, Cumbia, Paintball, Dodgeball, Baking cookies, Playing saxophone, Futsal, Making a sandwich, Riding bumper cars, Rope skipping, Removing ice from car, Long jump, Welding, Clean and jerk, Curling, Using the pommel horse, Mooping floor, Drinking coffee, Playing rubik cube, Playing water polo, Wrapping presents, Skateboarding, Removing curlers, Painting fence, Preparing pasta, Having an ice cream, Doing crunches, Grooming horse, Trimming branches or hedges, Shoveling snow, Kite flying, Playing violin, Blow-drying hair, Shuffleboard, Playing bagpipes, Playing drums, Polishing shoes, Playing piano, Playing polo, Blowing leaves, Archery, Brushing hair\\
		\hline
		\textbf{Novel:} $C_{100-120}$&
		Hurling, Laying tile, Rock climbing, Doing a powerbomb, Throwing darts, Using the balance beam, Playing field hockey, Elliptical trainer, Raking leaves, Painting furniture, Capoeira, Playing ice hockey, Snowboarding, Hand car wash, Baton twirling, Changing car wheel, Playing lacrosse, Playing beach volleyball, Tai chi, Knitting\\
		\hline
		\textbf{Novel:}$C_{120-200}$&
		Shaving, Gargling mouthwash, Getting a haircut, Vacuuming floor, Making a cake, Triple jump, Discus throw, Painting, Swinging at the playground, Ballet, Using the rowing machine, Shaving legs, Washing face, Braiding hair, Kneeling, Peeling potatoes, Snow tubing, Calf roping, Playing badminton, Tango, Bungee jumping, Playing accordion, Fun sliding down, Tumbling, Making a lemonade, Volleyball, Spread mulch, Rollerblading, Fixing bicycle, Drum corps, Pole vault, Cleaning shoes, Beach soccer, Windsurfing, Powerbocking, Croquet, Ironing clothes, Starting a campfire, Walking the dog, Getting a tattoo, Plataform diving, Sharpening knives, Tug of war, Hanging wallpaper, Surfing, Mowing the lawn, Assembling bicycle, Wakeboarding, Cleaning sink, Doing karate, Building sandcastles, Carving jack-o-lanterns, Waterskiing, Applying sunscreen, Brushing teeth, Playing squash, Playing pool, Scuba diving, BMX, Swimming, Decorating the Christmas tree, Longboarding, Sailing, Using uneven bars, Playing guitarra, Mixing drinks, Playing flauta, Drinking beer, Hand washing clothes, Playing kickball, Playing ten pins, Playing blackjack, Canoeing, Breakdancing, Shot put, Plastering, Beer pong, Clipping cat claws, Putting on makeup, Making an omelette\\
		\bottomrule
	\end{tabularx}
\end{table*}

\end{document}